# Supervised learning model for parsing Arabic language

[1]Nabil Khoufi, [2]Souhir Louati, [1]Chafik Aloulou and [1]Lamia Hadrich Belguith

ANLP Research Group-MIRACL Laboratory, University of Sfax, Tunisia
```
1{nabil.khoufi, chafik.aloulou, l.belguith}@fsegs.rnu.tn,
              2louati.sou@gmail.com
```

**Abstract.** Parsing the Arabic language is a difficult task given the specificities of this language and given the scarcity of digital resources (grammars and annotated corpora). In this paper, we suggest a method for Arabic parsing based on supervised machine learning. We used the SVMs algorithm to select the syntactic labels of the sentence. Furthermore, we evaluated our parser following the cross validation method by using the Penn Arabic Treebank. The obtained results are very encouraging.

## 1 Introduction

Within the Natural Language Processing (NLP) field, syntax represents the set of rules and the knowledge required to describe the logical sequence of the different lexical units of a particular sentence. Syntactic parsing represents an important step in the automatic processing of any language as it ensures the crucial task of identifying the syntactic structures of the sentences in a particular text. Several studies have been conducted in order to solve the problems of parsing. These efforts can be classified in three distinct approaches: the linguistic approach, the statistical approach, and the mixed or hybrid approach. The linguistic approach uses lexical knowledge and language rules in order to parse sentences whereas statistical approaches are essentially based on statistics or on probabilistic models. This type of approach is mainly based on the frequencies of occurrence that are automatically derived from the corpora. Last but not least, the hybrid approach is a mixture of the two previous ones: it integrates a linguistic analysis with a statistical one.

This paper is organized into four sections: in section 1, we present the works related to Arabic language parsing. Section 2 describes the different phases of the suggested method. Section 3 presents the principles and results of the evaluation. And section 4 presents our conclusions and suggestions for further research perspectives.

## 2 Related works

Many studies have focused on Arabic syntactic parsing. However, the number of these papers is very limited compared to the number of works dealing with other natural languages such as English or French. To our knowledge, the majority of efforts around Arabic language parsing use the symbolic approach based on rules. The latter gives satisfying results, but these are not yet at the English state-of-the-art level. (Ouersighni et al. 2001) developed a morphosyntactic analyser in modular form for Arabic. The analysis is based on the grammatical AGFL (Affixs Grammars over Finite Lattice) formalism. The analyser of (Othman et al. 2003) was realised in a modular form too and is based on the rules of the UBG (Unification Based Grammar) formalism. (Zemerli et al. 2004) have established a simple morphosyntactic analyser through the development of an application for vocalic synthesis of the Arabic language based on vowelized Arabic texts. This morphosyntactic analyser consists of two parts: the lexical database and the analysis procedure. In this analysis, the processing order of the text's words is crucial since it allows minimizing labelling errors. Aloulou (Aloulou2005) has developed a parsing system called MASPAR (Multi-Agent System for Parsing Arabic) based on a multi-agent approach. The chosen grammatical formalism is HPSG (Head-Driven Phrase Structure Grammar). It is a representation that permits to minimize the number of syntactic rules and to provide rich and well-structured lexical representations. The (Bataineh et al., 2009) analyser uses recursive transition networks. (Al-Taani et al. 2012) constructed a grammar under the CFG formalism (Context Free Grammar) and then implemented it in a parser with a top-down analysis strategy. All of these use hand-crafted grammars, which are time-consuming to produce and difficult to scale to unrestricted data. Moreover, these grammars do not fully cover all the specificities of the described language.

There are other Arabic parsers designed according to the statistical approach, which is based on statistical calculations or supervised learning techniques. (Tounsi et al., 2009) have developed a parser that learns from the treebank PATB (Penn Treebank Arabic) the functional labels in order to assign the respective syntactic structures to the different phrases according to the LFG (Lexical Functional Grammar) formalism. As an example, the analyser of (Ben Fraj 2010) learns from a corpus of syntactic tree patterns how to assign the most appropriate parse tree for syntactic interpretation of a new sentence. (Diab et al., 2007) present knowledge- and machine learning-based methods for tokenisation and basic POS tagging with a reduced tagset and base phrase chunking.

The study of related works shows that statistical methods for parsing the Arabic language remain largely untapped. It is difficult to compare the results because each parser uses a different evaluation metric. But according to the overview of the results of existing parsers, statistical-based parsers give better results than knowledge-based ones and are tested on a larger scale (see Table 1). These good results depend on the use of large amounts of annotated corpora. Since we have access to the ATB corpus and assume that the statistical analysers provide better results also with other languages (Charniak E. Et al., 2005 ) (Vanrullen T. Et al., 2006 ), we opted for a statistical method to build our annotation system of Arabic texts. More precisely, we use Machine learning techniques based on supervised learning. Table 1 present a comparison of evaluation results of parsers for the Arabic language.

**Table 1.** Comparison of the evaluation results

| System | Testing data | Results |
|---|---|---|
| Al-Taani et al. 2012 | 70 sentences | Accuracy 94 % |
| Bataineh et al. 2009 | 90 sentences | 85.6% correct<br>2,2% wrong<br>14,4% rejected |
| Mona Diab 2007 | ATB, 10 % | F-score 96.33%. |
| Ben Fraj 2010 | - | Accuracy 89,85 |

## 3 The suggested method

This section is devoted to the presentation of the general architecture of our suggested method.

The suggested method for parsing the Arabic language has two phases: the learning phase and the analysis phase. The first phase requires a training corpus, extraction features and a set of rules extracted from the learning corpus. The second phase implements the learning results from the first phase to achieve parsing. The phases of our approach are illustrated in the following figure:

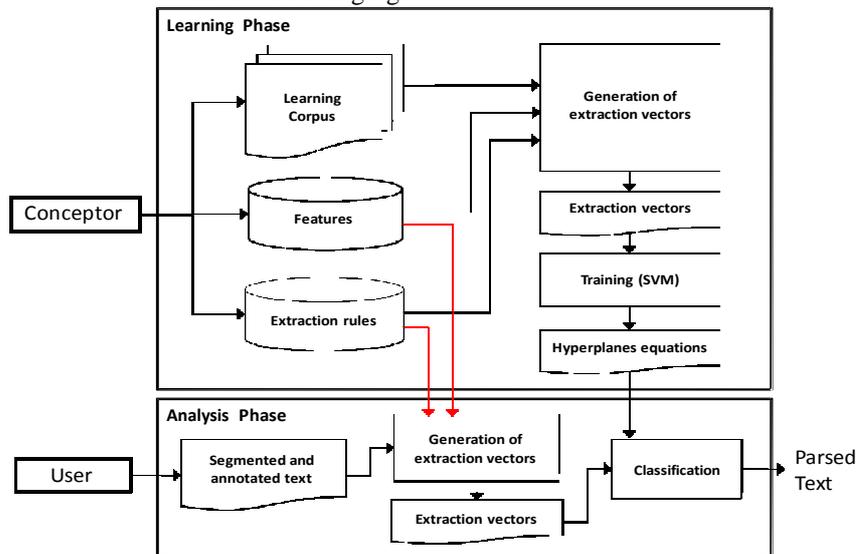

**Fig. 1.** The suggested method

### 3.1 The learning phase

The learning phase involves the use of a training corpus, a set of features and rules extracted from the learning corpus analysis in order to train the SVM (Support Vector Machine) classifier.

**Learning corpus.** The Penn Arabic Treebank (ATB) was developed in the laboratory of Linguistic Data Consortium (LDC) at the University of Pennsylvania (Maamouri M. et al., 2004). It is composed of data from standard and modern linguistic sources written in Arabic. It comprises 599 texts of different stories and news from the Lebanese newspaper An-Nahar. The texts in the corpus do not contain any vowels as it is typically in use in most texts written in Arabic. In the learning phase, we use the version ATB v3.2 of this corpus.

**Extraction features.** These features indicate the information used from the annotated corpus during the training stage, which is the morphological annotation.

We classified these features into two classes namely, part of speech features and contextual features:

A part of speech (POS) feature specifies the morphological category of the word being processed.

A contextual feature indicates the POS of the words in the left vicinity of the word being analyzed with a maximum depth equal to four.

The following table shows the different features used and their explanations:

**Table 2.** List of utilised extraction features.

|  | feature name | Explanation |
|---|---|---|
| A Part of speech feature | POS-W | Extract the morphological annotation of the word i which being processed. |
| Contextual features | POS-LEFT-i+1 | Extract the morphological annotation of the word in the left vicinity at position i +1. |
|  | POS-LEFT-i+2 | Extract the morphological annotation of the word in the left vicinity at position i +2. |
|  | POS-LEFT-i+3 | Extract the morphological annotation of the word in the left vicinity at position i +3. |
|  | POS-LEFT-i+4 | Extract the morphological annotation of the word in the left vicinity at position i +4. |

**Extraction rules.** These rules are derived from a deep analysis of the ATB. They are used to train our system in grouping the sequences of labels that may belong to the same syntactic grouping and thus better define their borders. The combination of features and rules extracted from the training corpus allows allocating each word in the sentence to its most probable syntactic group and thus training our analyser to classify them automatically. These rules have the following structure:

$$\text{Rule} : \{M1, M2, M3, M4, M5\} \quad Ci$$

Where M1 through M5 represents the morphological category of the words in a given syntactic group and Ci represents the syntactic class of the group. A rule may be composed of one, two, three, four or five elements. We extracted 53 rules from the ATB. We used the same tag set of the ATBs to simplify the learning process. Here are some examples of extracted rules:

R1 : PREP,NOUN   PP
R2 : ADJ,CONJ,ADJ   ADJP
R3 : NOUN_PROP   NP
R4 : PREP,NOUN,POSS_PRON   PP

**Generation of the extraction vectors.** This step aims to annotate each word of a sentence in the learning corpus according to the different extraction features presented above. Extraction rules are also used to identify the syntactic class of the groups of words.

Each group of word is described by a vector called extraction vector. The nominal value for a given feature corresponds to the morphological annotation of the word (adjective, noun, verb, punctuation, etc.) according to features used. This vector is completed by the appropriate syntactic class (NP, VP, PP ...) selected from the syntactic annotations in the ATB corpus. The set of extraction vectors forms an input file for the learning stage. At the end of this process, the learning corpus is converted from its original format into a vector format and we obtain a tabular corpus which consists of a set of vectors separated by a return as shown in the following example:

Vector1 : PREP,NOUN,?,?,?, PP
Vector2 : ADJ,CONJ,ADJ,?,?ADJP
Vector3 : NOUN_PROP,?,?,?,?,NP

**Learning.** This stage uses the previously generated extraction vectors in order to produce equations known as hyperplanes equations. The learning algorithm used in this stage is the SVM algorithm. To our knowledge, there is no works using SVMs for parsing the Arabic language. We decide to use SVMs for learning to test the potential of SVMs in parsing the Arabic language.

Since SVMs are binary classifiers, we have to convert the multi-class problem into a number of binary-class problems. This algorithm generates several hyperplane equations which are used to classify the different word groups according to their appropriate syntactic class (NP, PP, VP, ADJP ...). The training stage generates 25 hyperplane equations. It is noteworthy that the learning stage is done only once and is only repeated in case we increase the size of the corpus, or change the type of corpus.

This step is performed using 80 % of the ATB and the Weka library ( Frank, E. & Witten, Ian H., 2005). This tool takes as input extraction vectors in the form of an ".arff" file and outputs hyperplane equations.

### 3.2  The analysis phase

This phase implements the results of the learning phase in order to parse a sentence. The user must provide a segmented and a morphologically annotated text as input to our system. This phase proceeds in two steps as follows:

Firstly, a pre-processing is applied to the input sentence. Indeed, we use features and rules to arrange words in groups following the vector format as presented in the learning stage. This pre-processing generates extraction vectors like those generated as input for the learning stage. The only difference is that these vectors do not contain the syntactic class. This information will be calculated by the SVM classifier.

Then, the extraction vectors generated in the first step and the hyperplane equations generated in the learning stage are provided as input to the classification module. Indeed, for each vector, we calculate a score using hyperplane equations. Each equa-tion discriminates between two syntactic classes (e.g. PRT/ADVP). So every vector will have 25 scores according to the number of equations. The score and its sign are used to identify the suitable syntactic class for the test vector.

At the end of this stage we obtain a parsed sentence in a tree form.

## 4   Results

The evaluation of our analyser is achieved following the cross-validation method using the Weka tool. To realise that, we divided the training corpus into two distinct parts, one for learning (80%) and one for the test (20%). The results are exposed in the table 2.

**Table3.** Evaluation results.

| Precision | Rappel | F-score |
|-----------|--------|---------|
| 78.12 %   | 73.24% | 75.37%  |

The obtained results are encouraging and represent a good start for the implementation of supervised learning for parsing the Arabic language.

We noticed that the analysis of short sentences (<=20 words) presents the highest measures of recall and precision. As the sentence gets longer, there will be a more complex calculation, which reduces system's performance. This is due to the fact that our system does not handle very complex syntactic structures.

We believe that these results can be improved. In fact, we think that we can improve the learning stage by adding other features besides the POS features. As example of additional features, we can incorporate lexical data (external dictionary) to identify multi-word expressions. We will explore the effects of the integration of phrase functions on learning phase. During the implementation of our system, we noticed that the bigger the number of rules is, the higher the recall and precision are high. So we believe that the enrichment of our database of rules can significantly improve the results. The addition of syntactic rules is a solution to analyse long sentences.

## 5   Conclusion and perspectives

In this paper we presented our approach for Arabic parsing based on supervised learning. We used support vector machines, and we obtained an f-score of 0.99. As a perspective, we plan to integrate an efficient morphological analyser such as MADA in our system in order to process plain text. We intend to add other features like "group function" which already exists in the ATB and lexical data from external resources may be integrated to identify multi-word expressions.